\pdfoutput=1

\documentclass[11pt]{article}

\usepackage[]{EMNLP2023}

\usepackage{times}
\usepackage{latexsym}
\usepackage{comment}
\usepackage[T1]{fontenc}
\usepackage{amssymb}
\usepackage{graphicx}
\usepackage{wrapfig}
\usepackage[linesnumbered,ruled,vlined]{algorithm2e}
\usepackage{pifont}
\newcommand{\cmark}{\ding{51}}%
\newcommand{\xmark}{\ding{55}}%
\usepackage{multirow}

\usepackage[utf8]{inputenc}

\usepackage{microtype}

\usepackage{inconsolata}

\usepackage[colorinlistoftodos]{todonotes}
\usepackage{caption}
\usepackage{subcaption}

\title{Sensi-BERT: Towards Sensitivity Driven Fine-Tuning for Parameter-Efficient BERT}

\author{Souvik Kundu\thanks{Authors have equal contribution.} , Sharath Nittur Sridhar$^{*}$,  Maciej Szankin, Sairam Sundaresan\\
  Intel Labs, San Diego, USA \\}

\begin{document}
\maketitle
\begin{abstract}
Large pre-trained language models have recently gained significant traction due to their improved performance on various down-stream tasks like text classification and question answering, requiring only few epochs of fine-tuning. However, their large model sizes often prohibit their applications on resource-constrained edge devices. Existing solutions of yielding parameter-efficient BERT models largely rely on compute-exhaustive training and fine-tuning. Moreover, they often rely on additional compute heavy models to mitigate the performance gap. In this paper, we present \textit{Sensi-BERT}, a sensitivity driven efficient fine-tuning of BERT models that can take an off-the-shelf pre-trained BERT model and yield highly parameter-efficient models for downstream tasks. In specific, we perform sensitivity analysis to rank each individual parameter tensor, that then is used to trim them accordingly during fine-tuning for a given parameter or FLOPs budget. Our experiments show the efficacy of Sensi-BERT across different downstream tasks including \texttt{MNLI}, \texttt{QQP}, \texttt{QNLI}, \texttt{SST-2} and \texttt{SQuAD}, showing better performance at similar or smaller parameter budget compared to various alternatives.
\end{abstract}
\section{Introduction}
Large transformer based language models such as BERT \cite{devlin2018bert},
RoBERTa \cite{liu2019roberta} and ALBERT \cite{lan2020albert} have been extremely successful on many non-trivial natural language processing (NLP) tasks like question answering \cite{rajpurkar2016squad} and text classification \cite{wang2019glue}. Recently, the advent of generative pre-trained (GPT) \cite{floridi2020gpt} and various other large language models \cite{touvron2023llama} has further pushed the boundary of applications with the pre-trained language models, ranging from AI for science to code generation \cite{roziere2023code}. These models are generally pre-trained on a large unlabeled text corpus followed by fine-tuning on a task-specific data set. However, while their large model size usually helps them provide state-of-the-art (SoTA) accuracy on the downstream tasks, it limits their application and deployment on compute constrained edge devices.

Previous work has focused on reducing the BERT model sizes via  
several techniques including distillation \cite{sanh2019distilbert, jiao2019tinybert, sajjad2020poor}, pruning \cite{sanh2020movement,chen2020lottery} and quantization \cite{Zafrir_2019}. Another line of work relied on careful model design via removal of layers \cite{sridhar2022trimbert} or neural architecture search (NAS) \cite{xu2021bert}.
However, a majority of these techniques in yielding reduced size models are  iterative in nature and often rely on an extremely compute and storage-heavy teacher model making the fine-tuning extremely costly. Additionally, many of these methods require pre-training from scratch, thus are unable to compute-save via utilizing the wide selection of the pre-trained model pool \cite{xu2021bert}. Moreover, recent privacy concerns \cite{kundu2023learning} has sparked the need for both on-device fine-tuning and inference, making  efficient fine-tuning need of the hour, that the existing methods often fail to provide. 
\begin{figure}[t!] 
    \centering
    \includegraphics[width=0.43\textwidth]{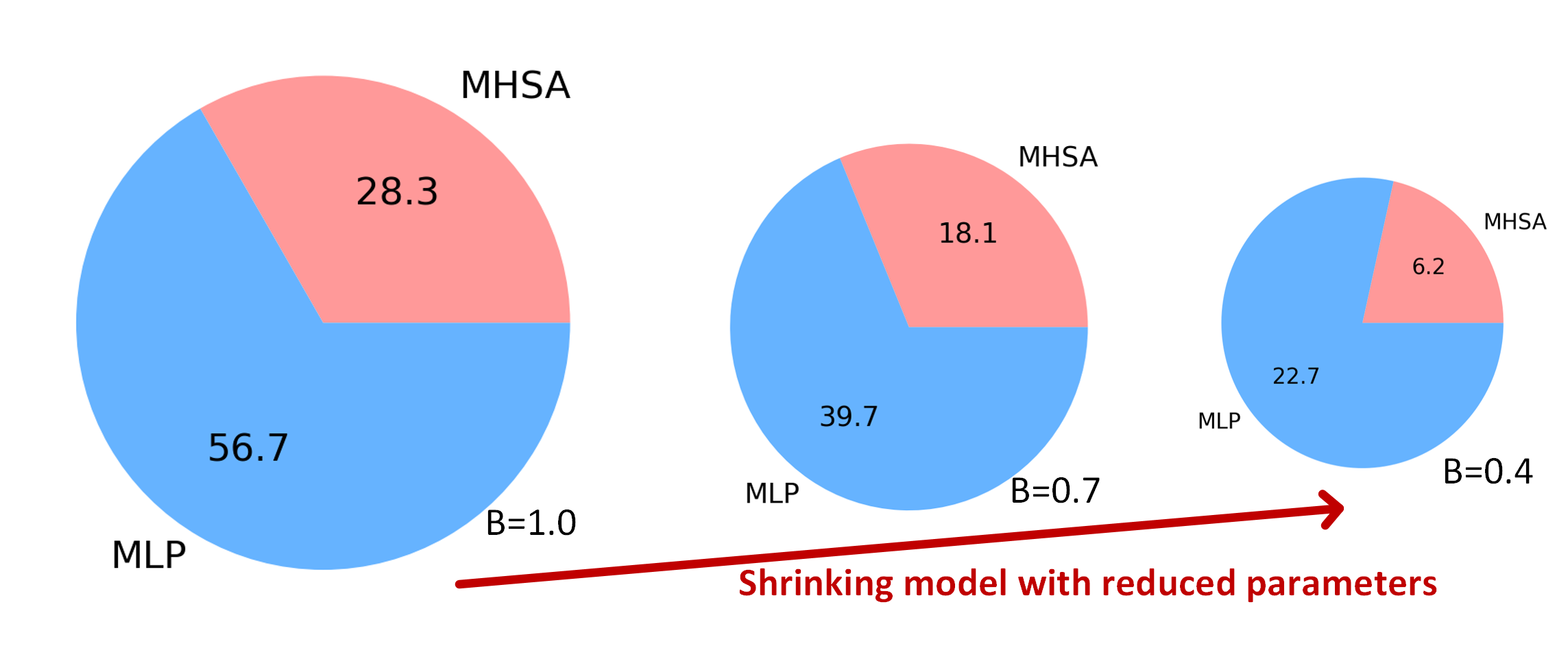}
    \vspace{-5mm}
    \caption{BERT-base MLP-MHSA parameter size distribution in million for different fine-tuning parameter budget in Sensi-BERT. Here, a budget $B$ corresponds to total non-zero parameter density $ \le B$. }
    \label{fig:sensibert_intro_compare}
    \vspace{-7mm}
\end{figure}

\textbf{Our Contributions.}
Towards the goal of providing BERT models for efficient fine-tuning as well as inference, we present \textit{Sensi-BERT}, a sensitivity-driven model trimming approach. In particular, starting from a pre-trained model, we first present a low-cost layer sensitivity analysis step to rank each of the intermediate dimensions of the self-attention and multi-layer perceptron (MLP) modules. We then, for a given parameter budget $B$, mask the low-important dimensions to zero and perform fine-tuning on the masked model only. Different from traditional pruning approach, that assigns a mask on each of the linear layers, we only assign it to the intermediate dimensions to get similar compression (see Fig. \ref{fig:sensibert_intro_compare}) at near-baseline accuracy. Note, here we directly meet the target budget without the need of any iterative tuning. Additionally, we performed analysis to leverage Sensi-BERT towards more efficient and less redundant model design. 

\section{Related Works}
Earlier research has focused on developing various small sized models \cite{xu2020bert, sun2019patient, xu2020bert, turc2019well} including DistilBERT \cite{sanh2019distilbert}, TinyBERT \cite{jiao2019tinybert}, and Poor Man's BERT \cite{sajjad2020poor}.  MobileBERT \cite{sun2020mobilebert} uses a bottom-to-top layer training with a custom loss to yield efficient models. However, these methods heavily rely on the knowledge distillation (KD) from a compute heavy teacher model that needs to be prepared during fine-tuning. Another area of research \cite{kurtic2022optimal, sanh2020movement, chen2020lottery} considered model compression during up-stream and down-stream training. However, apart from requirements of KD, many of these methods often require fine-tuning, compute, and storage of additional dense parameter tensors for significant epochs \cite{kurtic2022optimal, sanh2020movement}. Other methods require storage of initialized weights and iterative pruning \cite{chen2020lottery} even making  the fine-tuning costly. Recent works have also tried removing layers via careful manual effort \cite{sridhar2022trimbert} in identifying layer redundancy. Additionally, reduced model development via  NAS \cite{xu2021bert} has also been explored. However, they failed to utilize the pre-trained models available off-the-shelf. In contrast to these methods, we assume the fine-tuning forward compute budget to be same as the device's parameter budget. This poses a stricter constraint on both the fine-tuning and inference cost of the large models. Moreover, we also assume the use of a compute and memory heavy teacher model to be infeasible as often we prefer to perform both fine-tuning and inference at the resource-limited edge due to privacy issue \cite{kundu2021analyzing, kundu2023learning, zhang2023salvit}. Also, the use of teacher based distillation assumes the presence of a compute heavy model already fine-tuned on the downstream task, which may not be a practical scenario for many personalized down-stream datasets.

\section{Sensi-BERT: Methodology}
\label{sec:methodology}
Unlike many of the existing approaches, we take the advantage of pre-trained model weights and perform model size reduction during fine-tuning only. However, it is well known that efficient parameter reduction requires sensitivity-driven\footnote{Here sensitivity is analogous to layer importance; a layer with high sensitivity corresponds to more importance in retaining task performance and is computed by the proxy method of fraction of non-zero elements present for a given model parameter budget.} dropping of weights. Popular methods like compression via the alternating direction method of multipliers \cite{ren2019admm} assume that the sensitivity is hand-computed. Other methods use magnitude pruning \cite{chen2020lottery} on top of pre-trained weights. Here, we present a simple yet effective method of sensitivity evaluation using a pre-trained model.
\begin{figure}[t!] 

    \centering
    \includegraphics[width=0.47\textwidth]{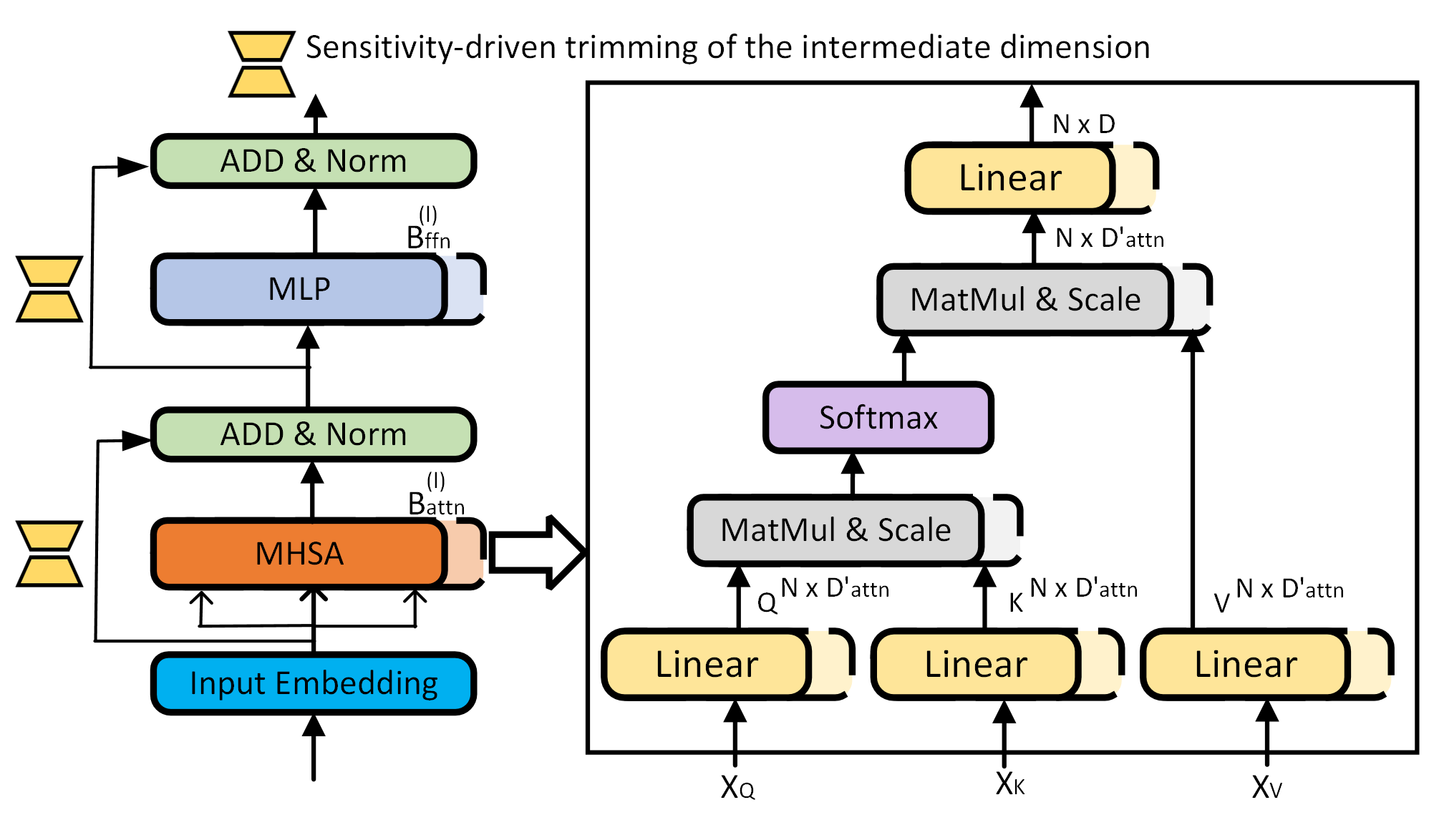}
    \vspace{-5mm}
    \caption{Sensi-BERT architecture overview. The solid portion of a block is representative of the trimmed down block-component.}
    \label{fig:sensibert_architecture}
    \vspace{-6mm}
\end{figure}
\subsection{Sensitivity Analysis}
Let a BERT model with $L$ layers each consisting of multi-head self attention (MHSA) followed by an MLP module, with each layer having $H$ heads. An MHSA module takes an input tensor $\textbf{X} \in \mathbb{R}^{N \times D_{in}}$ with sequence length and embedding dimension as $N$ and $D_{in}$, respectively. Each of the Query (Q), Key (K), and Value (V) linear transformation layers generates intermediate tensor $\textbf{T}_{mhsa} \in \mathbb{R}^{N \times D_{attn}}$ which finally gets projected to the output tensor $\textbf{O}_{mhsa} \in \mathbb{R}^{N \times D_{in}}$. For an MLP module the intermediate tensor size is $\textbf{T}_{mlp} \in \mathbb{R}^{N \times D_{ffn}}$ acting as the output and input of the $1^{st}$ and $2^{nd}$ fully connected (FC) layer, respectively, to finally produce output $\textbf{O}_{ffn} \in \mathbb{R}^{N \times D_{in}}$. Thus, it is evident that analysing the importance of the two intermediate tensor dimensions $D_{attn}$ and $D_{ffn}$ essentially translates to the sensitivity analysis of the MHSA and MLP modules for each layer. We thus define a set of learnable (non-binary) mask tensor $\mathbf{m}$, for each of MHSA and MLP intermediate tensor $\mathbf{m}_{mhsa} \in \mathbb{R}^{D_{attn}}$ and $\mathbf{m}_{mlp} \in \mathbb{R}^{D_{ffn}}$, respectively. The sensitivity analysis objective is 
\begin{figure*}[t!]
    \centering
    \includegraphics[width=0.99\textwidth]{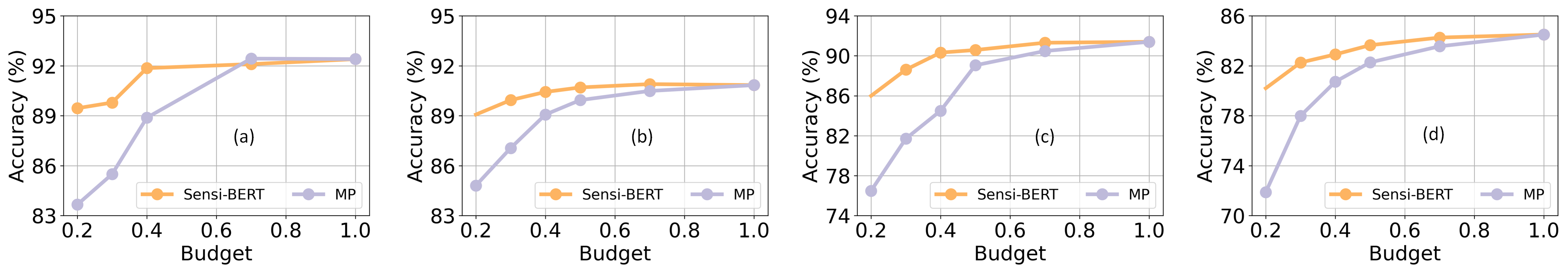}
    \vspace{-3mm}
    \caption{Comparison of Sensi-BERT with magnitude pruning (MP) on (a) \texttt{SST-2}, (b) \texttt{QQP}, (c) \texttt{QNLI}, and (d) \texttt{MNLI}.}
    \label{fig:sensi_vs_mp}
    \vspace{-3mm}
\end{figure*}
\begin{equation} 
    \min_{\Theta_{T}(f(\mathbf{X})),\mathbf{m}} \mathcal{L}_{CE}(\Phi(\mathbf{m}\odot\Theta_{T}(f(\mathbf{X})), \mathbf{y}) + ||\mathbf{m}||_{1}
\end{equation}
Here $f$, $\Theta_T$, and $\Phi$, represents the function generating query, key, and value (QKV) output, intermediate tensor, and the BERT function, respectively. 
Please note, during the sensitivity analysis process the mask tensor learns values corresponding to the importance of each dimension in $D_{attn}$ and $D_{ffn}$, for each head in each layer. We only allow the model to perform this optimization for one epoch thus minimizing the dense compute cost.

\subsection{Budgeted  Trimming During Fine-Tuning}
Once the mask tensor $m$ is trained, we provide the budget $B$ for the MHSA and MLP parameters. We then translate the budget to corresponding threshold set $\mathbf{m}_{th}$. We initialize an all 1s binary mask $\mathbf{b}$ and convert some of its location values to 0 by the following check
\begin{equation}
 \mathbf{b} = 0, \mathbf{m} \le \mathbf{m}_{th}
\end{equation}

For a model with total $d$ intermediate elements, we then apply the binary mask $\mathbf{b} \in \{0, 1\}^d$ to the model's intermediate tensor ensuring the model's parameters follow the set non-zero budget. Thus the fine-tuning objective becomes,
\begin{equation}
      \min_{\Theta_{T}(f(\mathbf{X}))} \mathcal{L}_{CE}(\Phi(\mathbf{b}\odot\Theta_{T}(f(\mathbf{X})), \mathbf{y}), ||\mathbf{b}||_0 \le B \times d  
\end{equation} 

Note, the fine-tuning requires only a fixed fraction of weights to be updated to non-zero. Thus the gradients associated with zero weights can be skip-computed, allowing a potential saving on the back propagation computation cost as well. Unless otherwise stated, we keep the budget for both the MHSA and MLP modules same in our evaluations. This ensures FLOPs reduction of similar proportion as the budget, due to linear relation between parameters and FLOPs for linear layers. Fig. \ref{fig:sensibert_architecture} depicts the architectural overview of Sensi-BERT. Note, the budget driven post thresholding of the binary mask $\mathbf{b}$, creates different budget for different layers driven by sensitivity. The figure shows $B^{(l)}_{attn}$ and $B^{(l)}_{ffn}$ to be the assigned non-zero intermediate tensor budget for the MHSA and MLP layer respectively for the $l^{th}$ layer.
\section{Experimental Evaluations}
\subsection{Models and Datasets} We use BERT-base \texttt{uncased} model to evaluate the performance of Sensi-BERT on four popular GLUE datasets, namely
 \texttt{QQP} - used for question similarity tasks in NLP and information retrieval;
\texttt{MNLI} \cite{williams2017broad} - crowdsourced, large-scale dataset for determining entailment, contradiction or neutrality in sentence pairs;
 \texttt{QNLI} \cite{wang2018glue} - it facilitates natural language inference tasks;
and  \texttt{SST-2} \cite{socher2013recursive} - a dataset for sentiment analysis using movie review sentences.  We compare our results with various popular baselines including DistilBERT. Benchmarks like TinyBERT and MobileBERT are excluded from our comparison as their loss is not architecture agnostic and they depend on additional data augmentation apart from KD \cite{xu2020bert}. Additionally, we demonstrate the performance of Sensi-BERT on a more complex dataset, namely the Stanford Question Answering Dataset (SQuAD) v1.1 \cite{rajpurkar2016squad}.

\begin{table*}[!t]
\small\addtolength{\tabcolsep}{-0pt}
\centering
\begin{tabular}{c|l|l|l|l|l|l|l}
\hline
{Method} & {\# Params}$\downarrow$ & {\texttt{MNLI}$\uparrow$} & {\texttt{QNLI}$\uparrow$} & {\texttt{QQP}$\uparrow$} & {\texttt{SST-2}$\uparrow$} & FD  & CPT\\ \hline
BERT-base    & 110 M  & 84.3 & 91.4 & 90.8 & 92.4 & -- & --                        \\ \hline
DistilBERT \cite{sanh2019distilbert}   & 66 M & 82.2 & 89.2 & 88.5 & 91.3 &  \cmark  & \xmark  \\  
BERT-PKD \cite{sun2019patient} & 66 M & 81.3  & 88.4  & 88.4  & 91.3 & \cmark  & \xmark \\
PD-BERT \cite{turc2019well}    & 66 M & \textbf{83.0} & 89.0 & 89.1 & 91.1&  \cmark  & \xmark  \\
Poor-Man's BERT \cite{sajjad2020poor}    & 66 M & 81.1 & 87.6 & 90.4 & 90.3&  \xmark  & \xmark  \\
Trim-BERT$_{n=4}$ \cite{sridhar2022trimbert}   & 67.6 M & 81.2 & 89.4 & 90.4 & 90.3 &  \xmark & \cmark  \\
NAS-BERT* \cite{xu2021bert} & 60 M & 83.3 & 91.3 & 90.9 & 92.0 &  \cmark  & \cmark  \\
BERT-of-Theseus \cite{xu2020bert}   & 66 M & 82.3 & 89.5  & 89.6 & 91.5 &  \xmark  & \xmark   \\  
Sensi-BERT$_{0.4}$ (Ours) & \textbf{53.3 M}   & 82.9 & \textbf{90.3} & \textbf{90.4} & \textbf{91.9} &  \xmark  & \xmark  \\ \hline
\end{tabular}
\caption{Performance of Sensi-BERT and other methods on various datasets. ``FD" and ``CPT" indicates fine-tuning distillation, and complex pre-training, respectively. Here, the subscript value of Sensi-BERT represents the budget. * = Representative results utilizing both FD and CPT.}
\vspace{-2mm}
\label{table:comparison_perf_table}
\end{table*}

\subsection{Results and Analysis}
 Following our methodology outlined in Section \ref{sec:methodology}, we performed sensitivity analysis for only one epoch with a batch size of 32 on each dataset. With the dimensions ranked,  we apply thresholding to create the binary mask for any given budget. Note, we perform the sensitivity analysis \textbf{only once} and perform fine-tuning once for just three epochs for each budget. So, to generate $N$ fine-tuned models of $N$ different budgets the total cost is $1 + 3N$. Fig. \ref{fig:sensi_vs_mp} shows results of Sensi-BERT at different parameter budget. We also compare the results with standard magnitude pruning (MP), wherein we apply the sparse mask to the attention tensors based on top-k magnitude in the query parameter tensor. For the the MLP layers we select top-k magnitude locations of the FC layer weights. Upon creating the binary sparse mask based on the top-k locations post pre-training, we keep the mask frozen during fine-tuning to have a fair comparison with us.

 \textbf{Observation.} \textit{Magnitude pruned models yield similar accuracy as Sensi-BERT at high budget while yield significantly poorer accuracy at the lower parameter regime.}

As Fig. \ref{fig:sensi_vs_mp} shows that Sensi-BERT yields significant better accuracy by up to $9.5\%$ (\texttt{QNLI} at $B=0.2$) compared to MP, clearly highlighting the importance of the sensitivity analysis step. However, for higher $B$, the model may still remain over-parameterized highlighting the reduced importance of sensitivity driven model trimming.

\subsection{Comparison with Alternatives}
We compare our approach with various reduced parameter BERT design methodologies in Table \ref{table:comparison_perf_table}. As we can see, despite not leveraging complex fine-tuning models for distillation Sensi-BERT yields similar and often better accuracy than these methods. Although NAS-BERT leverage more complex and compute heavy pre-training and fine-tuning along with pre-training KD and data augmentation, we place the results in the table as a representative to demonstrate the closeness of performance of models yielded via Sensi-BERT. BERT-of-Theseus \cite{xu2020bert} follows similar philosophy as ours and only use the CE loss for fine-tuning. 
It is noteworthy that, our method is orthogonal to most of these approaches and can also be deployed with these methods during final fine-tuning steps. 

\subsection{Results on SQuAD v1.1}
\begin{figure}[!h]
    \centering
    \includegraphics[width=0.48\textwidth]{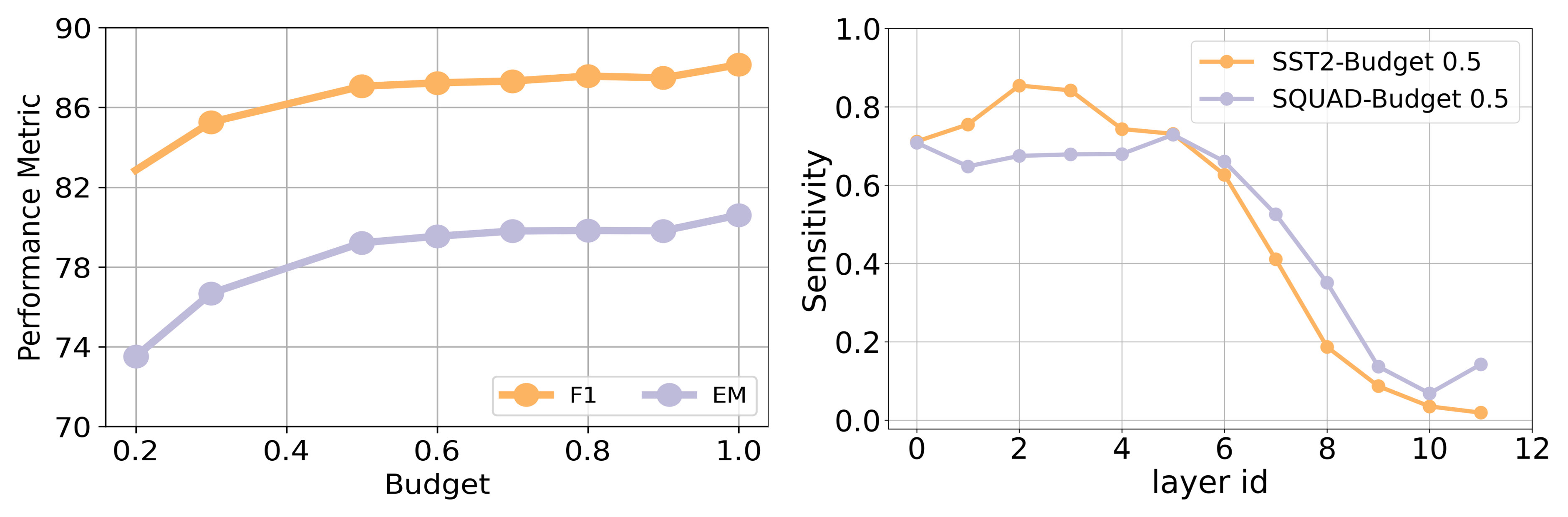}
    \vspace{-4mm}
    \caption{(a) Performance of Sensi-BERT on SQuAD v1.1, (b) Sensitivity comparison of MLP modules for SQuAD v1.1 and SST-2 GLUE task.}
    \label{fig:acc_f1_em_squad}
    \vspace{-3mm}
\end{figure}

 SQuAD is a popular reading comprehension dataset, used to evaluate a model’s ability to read a passage of text and then answer questions about it.
 We use the F1-score and the exact match (EM) score to evaluate Sensi-BERT's performance on SQuAD v1.1 on various budgets. We performed one epoch of sensitivity analysis, followed by two epochs of fine-tuning. As shown in Fig. \ref{fig:acc_f1_em_squad}(a), the model's F1 and EM score remains close to that with the dense model even at a budget of 0.5. This clearly demonstrates the effectiveness of sensitivity driven resource-limited fine-tuning even for complex tasks like SQuAD.
 
\textbf{Observation.} \textit{Layer sensitivity follows similar trend across two different tasks.}

As Fig. \ref{fig:acc_f1_em_squad}(b) shows the MLP layer sensitivity for two different NLP tasks for same target parameter budget of $0.5$. Interestingly, despite significant difference between the task types and complexity, the sensitivity follows similar trend. This opens up an interesting question of whether \textit{sensitivity can be considered transferable} across tasks.

\subsection{Sensitivity-Driven Architecture Analysis}
\begin{figure}[!t]
    \centering
    \includegraphics[width=0.49\textwidth]{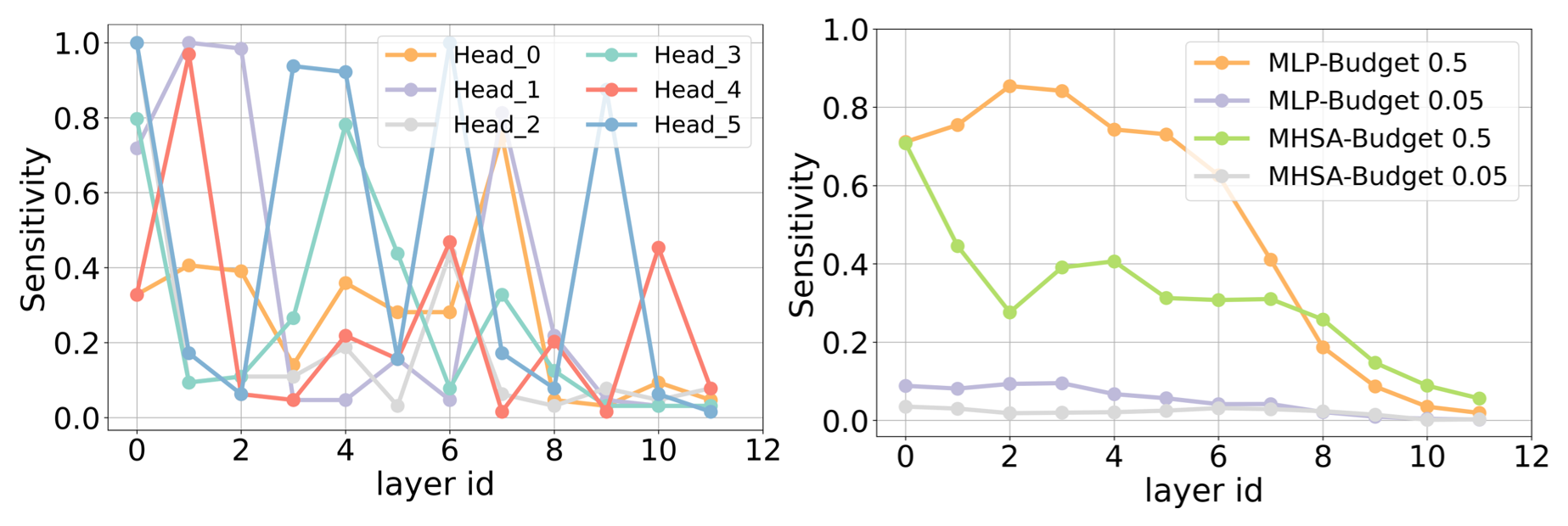}
    \vspace{-2mm}
    \caption{(a) MHSA head-wise sensitivity distribution across layers, for the first six heads; (b) Sensitivity of the MHSA and MLP modules layer-wise for different parameter budgets.}
    \label{fig:senstivity_analysis}
    \vspace{-3mm}
\end{figure}
\textbf{Observation.} \textit{While different MHSA heads show different sensitivity trend, the MHSA and MLP layer sensitivity consistently reduces as we go deeper in the model.}

Fig. \ref{fig:senstivity_analysis}(a) shows the results of head-wise sensitivity of BERT-base for $B=0.5$ on \texttt{SST-2}. It is clear that while some heads carry more sensitivity, few other heads hint to be more over-parameterized (example, head 2 and 3 in the Fig. \ref{fig:senstivity_analysis}(a)). Fig. \ref{fig:senstivity_analysis}(b), on the contrary, shows a clear decreasing sensitivity trend for both MHSA and MLP modules. Using this findings, we designed a custom BERT having reduced intermediate dimensions $D_{ffn}$ at its last three layers. The goal is to check whether we can leverage this findings in designing more compact models that can yield similar accuracy. As shown in the Table \ref{table:architec_analysis_ablation}, the models with reduce $D_{ffn}$ provide similar or better accuracy highlighting the utility of Sensi-BERT as a tool to guide architecture design. 
\begin{table}[!h]
\small\addtolength{\tabcolsep}{-0pt}
\centering
\begin{tabular}{c|l|l|l}
\hline
\multicolumn{4}{c}{{\texttt{SST-2} Accuracy (\%)}} \\ \hline
{$D_{ffn}$: 3072} & {$D_{ffn}$: 2048} & {$D_{ffn}$: 1024} & {$D_{ffn}$: 512}\\ \hline
92.4 & 92.3 & \textbf{92.7} & 92.0 \\ \hline
\end{tabular}
\caption{Performance of a BERT-base with various intermediate dimension $D_{ffn}$ for the MLP module at the later layers (in specific, at $10^{th}$, $11^{th}$, and $12^{th}$ layer).}
\vspace{-4mm}
\label{table:architec_analysis_ablation}
\end{table}
\section{Conclusions and Future Work}
In this paper we presented Sensi-BERT, a sensitivity-driven approach in yielding parameter-efficient BERT for efficient fine-tuning and inference. We leveraged layer-wise sensitivity of the intermediate tensors to identify layer-importance and perform fine-tuning on only fraction of model parameters evaluated through this sensitivity. 
Compared to various alternatives leveraging compute heavy models during fine-tuning Sensi-BERT demonstrated to yield similar or improved performance at a much less compute and storage demand. Leveraging such sensitivity driven approach for large foundation models in reducing their compute footprint is an interesting future research.
\bibliography{main}
\bibliographystyle{acl_natbib}

\end{document}